\documentclass[12pt,twoside,a4paper]{article}
\usepackage{amsmath,amssymb,latexsym,theorem,natbib,epsfig,color,subfigure}
\usepackage{multirow,graphicx,array,rotating}  
\usepackage{epstopdf,url,afterpage}
\newfont{\msa}{msam10 scaled\magstep1}
\newfont{\ssmsa}{msam9}

\def\crps{\mathop{\hbox{\rm CRPS}}}
\def\crpss{\mathop{\hbox{\rm CRPSS}}}
\def\logs{\mathop{\hbox{\rm LogS}}}
\def\logss{\mathop{\hbox{\rm LogSS}}}
\def\sign{{\mathrm {sign}}}

\numberwithin{equation}{section}

\title{Machine learning for total cloud cover prediction}

\author{{\sc \'Agnes Baran$^1$}, {\sc Sebastian Lerch$^2$}, {\sc Mehrez El Ayari$^{1,3}$}\\ and {\sc S\'andor Baran$^1$} \vspace*{0.5cm}\\
         $^1$Faculty of Informatics, University of Debrecen\\
         Kassai \'ut 26, H-4028 Debrecen, Hungary\\
         $^2$Institute for Stochastics, Karlsruhe Institute of Technology, \\
         Englerstra\ss{}e 2, D-76128 Karlsruhe, Germany \\
         $^3$Doctoral School of Informatics, University of Debrecen\\
         Kassai \'ut 26, H-4028 Debrecen, Hungary 
        }

\date{}
\begin{document}
\pagestyle{myheadings}

\maketitle

\begin{abstract}
Accurate and reliable forecasting of total cloud cover (TCC) is vital for many areas such as astronomy, energy demand and production, or agriculture. Most meteorological centres issue ensemble forecasts of TCC, however, these forecasts are often uncalibrated and exhibit worse forecast skill than ensemble forecasts of other weather variables. Hence, some form of post-processing is strongly required to improve predictive performance. As TCC observations are usually reported on a discrete scale taking just nine different values called oktas, statistical calibration of TCC ensemble forecasts can be considered a classification problem with outputs given by the probabilities of the oktas. This is a classical area where machine learning methods are applied. We investigate the performance of post-processing using multilayer perceptron (MLP) neural networks, gradient boosting machines (GBM) and random forest (RF) methods. Based on the European Centre for Medium-Range Weather Forecasts global TCC ensemble forecasts for 2002--2014 we compare these approaches with the proportional odds logistic regression (POLR) and multiclass logistic regression (MLR) models, as well as the raw TCC ensemble forecasts. We further assess whether improvements in forecast skill can be obtained by incorporating ensemble forecasts of precipitation as additional predictor. Compared to the raw ensemble, all calibration methods result in a significant improvement in forecast skill. RF models provide the smallest increase in predictive performance, while MLP, POLR and GBM approaches perform best. The use of precipitation forecast data leads to further improvements in forecast skill and except for very short lead times the extended MLP model shows the best overall performance.

\bigskip
\noindent {\em Key words:\/} ensemble calibration; gradient boosting machine; logistic regression; multilayer perceptron; random forest; total cloud cover
\end{abstract}

\section{Introduction}
\label{sec1}

Reliable and accurate prediction of total cloud cover (TCC) has a principal importance in observational astronomy \citep{yc13} and in the prediction of photovoltaic energy production, as it is the main cause of variation in solar-radiation energy supply \citep{mat12,mmc12}, but it is also of great relevance in agriculture, tourism and in some other fields of economy. According to the definition of the World Meteorological Organization ``total cloud cover is the fraction of the sky covered by all the visible clouds'' \citep{wmocloud}. Even though this definition indicates a continuous quantity in the \ $[0,1]$ \ interval, TCC observations are usually reported in eighths of sky cover called oktas taking just nine different values. In this way TCC forecasting can be considered as a nine-group classification problem and thus requires markedly different methods than those used for other weather variables such as temperature, wind speed or precipitation accumulation, which are treated as continuous quantities.

TCC forecasts are generated using numerical weather prediction (NWP) models \citep[for a comparison of the performance of the state of the art techniques see e.g.][]{kcbhv19}, and recently all major meteorological centres issue ensemble forecasts of TCC using their operational ensemble prediction systems (EPSs). Examples include the Global Ensemble Forecast System of National Centers for Environmental Prediction \citep{zzhlpw17} or the EPS of the independent intergovernmental European Centre for Medium-Range Weather Forecasts \citep[ECMWF;][]{mbpp96,lp08,ecmwf12}. However, ensemble forecasts often tend to be underdispersive, that is the spread of the ensemble is too small to accurately capture the full uncertainty, and can be subject to systematic bias. This phenomenon can be observed with several operational EPSs \citep[see e.g.][]{bhtp05,pbl08} calling for some form of statistical post-processing \citep{buizza18}. TCC ensemble forecasts are even more problematic, as in terms of forecast skill they highly underperform ensemble forecasts of other weather variables such as temperature, wind speed, pressure or precipitation \citep[see e.g.][]{hfab15,ecmwfEval18}.

Over the past decade, various statistical post-processing methods have been proposed in the meteorological and statistical literature, for an overview see e.g.\ \citet{wfk14} or \citet{vwm18}. These include parametric approaches like Bayesian model averaging \citep{rgbp05} or non-homogeneous regression \citep{grwg05} providing estimates of the probability distributions of the weather quantities of interest, non-parametric techniques like quantile regression \citep[see e.g.][]{fh07,brem19} or mixed methods such as quantile mapping \citep[see e.g.][]{hs18,glhrblp19}. Recently machine learning methods have become more and more popular in ensemble post-processing. For example, \citet{tmzn16} used quantile regression forests (QRF) for calibration of ensemble forecasts of temperature and wind speed, and \citet{tfnm19} recently extended the technique to precipitation forecasts. \citet{rl18} applied neural networks for post-processing of ECMWF near-surface temperature ensemble forecasts using QRF as a benchmark model, whereas \citet{brem20} employed neural networks in quantile function regression for calibrating ensemble forecasts of wind speed. \citet{bwks19} compare several machine learning approaches for post-processing NWP predictions of solar radiation based on quantile regression, including random forests, gradient boosting and neural networks.

The discrete nature of TCC means that the predictive distribution should take the form of a discrete probability distribution and post-processing can be considered as a classification problem resulting in the probabilities of the oktas. For calibrating TCC ensemble forecasts \citet{hhp16} propose two discrete parametric post-processing approaches, namely multiclass logistic regression \citep[MLR;][]{izen08} and proportional odds (or ordered) logistic regression \citep[POLR;][]{mccull80}. Different versions of logistic regression had already been successfully applied in statistical post-processing \citep[see e.g.][]{wilks09,sk10} and ordered logistic regression also showed good performance for forecasts of discrete categories \citep{mmwz14}. 

Since probabilistic multi-category classification is one of the main areas of application of machine learning, the main goal of our work here is to investigate the use of machine learning methods for total cloud cover prediction in the framework of statistical post-processing of TCC ensemble forecasts. With the help of ECMWF global ensemble forecasts for the period 2002 -- 2014 we test the performance of multilayer perceptron neural networks \citep[MLP;][]{dlbook}, gradient boosting machine \citep[GBM;][]{f01} and random forest algorithms \citep[RF;][]{breiman01}, and compare their forecast skill with the raw TCC ensemble and the MLR and POLR approaches of \citet{hhp16}. We further investigate the effect of using precipitation ensemble forecasts as additional predictors in TCC post-processing.

The paper is organized as follows. Section \ref{sec2} contains the description of the TCC and precipitation ensemble forecasts and observations. Section \ref{sec3} reviews the various calibration methods and tools used for forecast evaluation. A case study on post-processing of TCC ensemble forecasts is provided in Section \ref{sec4}, and the article concludes with a discussion in Section \ref{sec5}. 

\section{Data}
\label{sec2}

We consider 52-member ECMWF global ensemble forecasts (high-resolution forecast (HRES), control forecast (CTRL) and 50 members (ENS) generated using random perturbations) of TCC and 24 h precipitation accumulation initialized at 1200 UTC for 10 different lead times ranging from 1 day to 10 days for the period between 1 January 2002 and 20 March 2014, together with the corresponding observations. The TCC data set is identical to the one investigated in \citet{hhp16} containing data for 3330 synoptic observation (SYNOP) stations left after an initial quality control. TCC SYNOP observations are reported in values \ ${\mathcal Y}=\{0,0.1,0.25,0.4,0.5,0.6,0.75,0.9,1\}$ \ corresponding to the different oktas, whereas the raw ensemble forecasts are continuous values in the \ $[0,1]$ \ interval. The matching of forecasts and observations is performed with quantization of forecast values using intervals
\begin{align*}
  [0,0.01[, \quad &[0.01,0.1875[, \quad [0.1875,0.3125[, \quad [0.3125,0.4375[, \quad [0.4375,0.5625[, \\ &[0.5625,0.6875[, \quad [0.6875,0.8125[, \quad [0.8125, 0.99[, \quad [0.99,1],
  \end{align*}
  that is raw or post-processed forecasts falling e.g. into the interval \ $[0.1875,0.3125[$ \ correspond to observation value $0.25$ \citep[see][Table A1]{hhp16}.

Our additional precipitation data set, which has been investigated in \citet{hspbh14}, contains forecast-observation pairs for 2917 SYNOP stations after quality control. At 2239 of these station both TCC and precipitation data are available.

\section{Calibration methods and forecast evaluation}
\label{sec3}

In what follows, let \ $Y\in {\mathcal Y}=\{y_1,y_2,\ldots ,y_9\}$ \ be TCC  at a given location and time expressed in oktas and denote by \ $\boldsymbol f = (f_1,f_2, \ldots ,f_{52})$ \ the corresponding 52-member ECMWF TCC ensemble forecast with a given lead time, where \ $f_1=f_{\text{HRES}}$ \ and  \ $f_2=f_{\text{CTRL}}$ \ are the high-resolution and control members, respectively, whereas \ $f_3,f_4, \ldots ,f_{52}$ \ correspond to the 50 statistically indistinguishable (and thus exchangeable) ensemble members \ $f_{\text{ENS},1},f_{\text{ENS},2}, \ldots ,f_{\text{ENS},50}$ \ generated using random perturbations. In this discrete setting the estimation of the predictive distribution of \ $Y$ \ reduces to the estimation of conditional probabilities
\begin{equation}
  \label{eq:condPMF}
  {\mathsf P}\big(Y=y_k \mid \boldsymbol f\big), \qquad k=1,2,\ldots ,9.
\end{equation}
Obviously, in \eqref{eq:condPMF} the raw ensemble forecast \ $\boldsymbol f$ \ can be replaced by any feature vector \ $\boldsymbol x$ \ derived from the ensemble and/or other covariates. In order to ensure comparability with the reference MLR and POLR approaches for classification using TCC data only (see Section \ref{subs4.1})  we consider the same feature set as in  \citet{hhp16}. The investigated covariates are the HRES forecast \ $f_{\text{HRES}}$,\  the control forecast \ $f_{\text{CTRL}}$, \ the mean of the 50 exchangeable ensemble members \ $\overline f_{\text{ENS}}$, \ the ensemble variance
$$s^2:=\frac 1{51} \sum_{i=1}^{52} \big(f_i - \overline f\big)^2, \qquad \text{where} \qquad \overline f :=\frac 1{52}\sum_{i=1}^{52}f_i,$$
the proportions of forecasts predicting zero and maximal cloud cover
$$p_0:=\frac 1{52} \sum_{i=1}^{52} {\mathbb I}_{\{f_i=0\}} \qquad \text{and} \qquad p_1:=\frac 1{52} \sum_{i=1}^{52} {\mathbb I}_{\{f_i=1\}},$$
respectively, where \ ${\mathbb I}_H$ \ denotes the indicator function of a set \ $H$, \ and an interaction term
$$I:=s^2 \sign(d)d^2 \qquad \text{with} \qquad d:=\big((f_{\text{HRES}}-0.5)+(f_{\text{CTRL}}-0.5)+(\overline f_{\text{ENS}}-0.5)\big)/3$$
connecting the ensemble variance and the mean deviation of the first three features from $0.5$.

As additional feature we also consider the mean \ $\overline f_{\text{PREC}}$ \ of the ECMWF 51-member precipitation ensemble forecast for some of the models (see Section \ref{subs4.2}). The use of the HRES precipitation forecast or of the mean of the 52-member precipitation ensemble (including HRES) instead of \ $\overline f_{\text{PREC}}$ \ was also tested, however, these models did not result in a significant improvement in the forecast skill.

In the following, we introduce the different post-processing models for TCC. Implementation details for all models are provided in Section \ref{subs4.1-impl}.

\subsection{Multiclass and proportional odds logistic regression}
\label{subs3.3}

In multiclass logistic regression, after choosing an arbitrary reference class, the log-odds of a remaining class with respect to the reference class is expressed as an affine function of the features.  This means that after setting  e.g.\ the last okta \ $y_9$ \ as reference class, the conditional distribution of TCC with respect to an $M$-dimensional feature vector \ $\boldsymbol x$ \ equals
\begin{equation}
  \label{eq:mlr}
  {\mathsf P}\big(Y=y_k \mid \boldsymbol x\big)=
  \begin{cases}
    \frac{{\mathrm e}^{L_k(\boldsymbol x)}}{1+\sum_{\ell=1}^8{\mathrm e}^{L_{\ell}(\boldsymbol x)}}, & k=1,2,\ldots ,8; \\
    \frac 1{1+\sum_{\ell=1}^8{\mathrm e}^{L_{\ell}(\boldsymbol x)}}, & k=9, 
  \end{cases}
  \quad \text{with} \quad L_k(\boldsymbol x):= \beta_{0k} + \boldsymbol x^{\top}\boldsymbol\beta_k,
\end{equation}
where \ $\beta_{0k} \in {\mathbb R}, \ \boldsymbol\beta_k \in {\mathbb R}^M$,  \ resulting in \ $8(M+1)$ \ free parameters to be estimated on the basis of the training data. 

The POLR model is designed to fit ordered data such as the TCC observations at hand. Given a feature vector \ $\boldsymbol x$, \ the conditional cumulative probabilities  of \ $Y$ \ are expressed as
\begin{equation}
  \label{eq:polr}
  {\mathsf P}\big(Y\leq y_k \mid \boldsymbol x\big)= \frac{{\mathrm e}^{\mathcal L_k(\boldsymbol x)}}{1+{\mathrm e}^{\mathcal L_k(\boldsymbol x)}},  \qquad \text{with} \qquad \mathcal L_k(\boldsymbol x):= \gamma_{0k} + \boldsymbol x^{\top}\boldsymbol\gamma, \quad \ k=1,2,\ldots, 9, 
\end{equation}
where we assume that \ $\gamma_{01}<\gamma_{02}< \cdots < \gamma_{09}$. \ In this way POLR model \eqref{eq:polr} is more parsimonious than MLR model \eqref{eq:mlr}, as it has just \ $9+M$ \ unknown parameters.

\subsection{Multilayer perceptron neural network}
\label{subs3.1}

A multilayer perceptron (MLP) is a classical feedforward neural network, consisting of an input layer, an output layer and some intermediate layers (so-called hidden layers) comprised of several neurons each. The value in each of the neurons is a transformed value (via an activation function) of a weighted sum of all neuron values from the previous layer plus a bias term. The number of neurons in the input and output layers are uniquely determined by the number of features and number of classes, respectively, whereas the number of the hidden layers and the number of the neurons in a particular hidden layer are free (or tuning) parameters of the network. For a comprehensive introduction to neural networks, see e.g. \citet{dlbook}.

The network is trained using a set of labeled data (training set): the weights of the neurons are determined in order to minimize a given loss function on the training set. To avoid overfitting it is recommended to use early stopping rules based on a validation set. Typically it is a randomly chosen subset of the labeled data set available for the training. The minimization process terminates if the value of the loss function computed on the validation set does not improve during a given number of subsequent iterations. Similar techniques are applied for the other machine learning methods, see Section \ref{subs4.1-impl} for details.

Another tool to prevent overfitting is the extension of the loss function with a regularization term. Here we use an \ $L_2$ \ regularization where the sum of squares of the weights of the network is multiplied by a factor (which is an additional tuning parameter of the network). The trained network provides for each feature vector a probability distribution corresponding to the oktas. 

\subsection{Random forest models and gradient boosting machines}
 \label{subs3.2}
 
Random forests (RF) and gradient boosting machines (GBM) are machine learning models which are both based on ensembles of decision trees. Decision trees are flowchart-like structures that have been used in meteorological forecasting since the 1950s \citep{mcg17}. Decision tree models are obtained through iteratively splitting training data into groups according to a threshold in one of the features \ $\boldsymbol{x}$ \ which is chosen to maximize the homogeneity of the target variable within the resulting subsets. This process is iterated until a stopping criterion is reached. Out-of-sample forecasts can be obtained by proceeding through the decision tree according to the predictor input, and estimating class probabilities by the the empirical frequencies of observed classes in the corresponding subset of the recursively partitioned feature space. While there exist several algorithms for decision tree learning, we will here focus on classification and regression trees (CART) first introduced by \citet{CART}. 

\subsubsection*{Random forest models}

To improve robustness and address overfitting issues of decision trees, random forest models \citep{breiman01} repeatedly resample the training set to obtain multiple decision trees. This bootstrap aggregation (or bagging) approach is used in conjunction with only considering a random subset of the predictors at each splitting node. Class probability predictions for out-of-sample cases are obtained by averaging over the decision trees in the RF ensemble.

Several tuning parameters have to be chosen when implementing RF models. Most importantly, the number of trees in the forest has to specified, and the depth (the number of levels of recursive partitioning) as well as the number of predictor variables randomly selected at each splitting node have to be selected for the individual trees. Generally, RF models are often relatively robust to these tuning parameters and tend to not be prone to overfitting for a wide range of parameter choices.

\subsubsection*{Gradient boosting machines}

In contrast to randomly resampling the training data, gradient boosting machines consist of ensembles of decision trees which are grown sequentially, using information from previously grown trees. Thereby, each decision tree is fit on a modified version of the original training set focusing on regions where previous model iterations provide poor predictions. 

The umbrella term boosting refers to machine learning algorithms that fit models by combining several simpler models, decision trees in our case. Following \citet{f01}, various notions of gradient boosting have been developed and it was demonstrated that boosting can be interpreted as gradient descent algorithm in function space where a loss function is iteratively optimized by choosing a function that points in the negative gradient direction. Gradient boosting principles are applicable for wide range of loss functions, and corresponding algorithms have been developed for a wide range of machine learning tasks. For a general introduction to gradient boosting see e.g.  \citet{htf09}.

We here employ a specific variant of tree-based gradient boosting called extreme gradient boosting \citep{xgboost}, which relies on second-order approximations of the objective function. GBM model predictions are obtained via
\begin{equation}
\hat z^c = \sum_{m=1}^M h_m^c(\boldsymbol{x}), \label{eq:gbm}
\end{equation}
where \ $h_m^c$ \ denotes a regression tree for category $\ c \in\{1,\dots,9\}\ $ containing a continuous value in all terminal leaves, and \ $M$ \ is the number of boosting iterations. For probabilistic classification tasks,  separate sets of regression trees are fitted simultaneously for all categories, and the  obtained latent values \ $\hat z^c$ \ are transformed according to a softmax function. A regularized version of the LogS (see Section \ref{subs3.4}) is used to learn the set of functions used in the model \eqref{eq:gbm}. For details, see \citet{xgboost}. 

Compared to RF models, GBM often provide better predictions in a variety of applications, but are more prone to overfitting and more difficult to tune. In particular, the number of boosting iterations, \ $M$, \ is of crucial importance. Further, the complexity of the individual trees \ $h_m$ \ must often be restricted, see Section \ref{subs4.1-impl} for details.

\subsection{Verification scores}
 \label{subs3.4}
 
As discussed in \citet{gbr07}, the main goal of probabilistic forecasting is to maximize the sharpness of the predictive distribution subject to calibration. Sharpness refers to the concentration of the predictive distribution, whereas calibration means a statistical consistency between forecasts and observations. These two goals can be simultaneously addressed with the help of proper scoring rules, which are loss functions \ $\mathcal S (F,x)$ \ assigning numerical values to pairs \ $(F,x)$ \ of forecasts and observations. As mentioned in the Introduction, in the case of TCC by forecast \ $F$ \ we refer to a discrete probability distribution on \ ${\mathcal Y}$ \ characterized by a probability mass function (PMF) \ $p_F(y)$.  

In the atmospheric sciences probably the most popular proper scoring rules are the logarithmic score \citep[LogS;][]{good52} and the continuous ranked probability score \citep[CRPS;][]{gr07,wilks11}. The former is the negative logarithm of the PMF evaluated at the observation, that is
\begin{equation*}
  \logs\big(F,x\big):= -\log \big(p_F(x)\big),
\end{equation*}
whereas for TCC probabilistic forecasts at hand the latter can be given as
\begin{equation*}
 \crps\big(F,x\big)=\sum_{k=1}^9 p_F(y_k) \big| y_k -x \big| - \sum_{k=2}^9 \sum_{\ell=1}^{k-1} p_F(y_k) p_F(y_{\ell}) \big| y_k - y_{\ell}\big|,
\end{equation*}
which is the discrete version of the representation
\begin{equation*}
 \crps\big(F,x\big)={\mathsf E}|X-x|-\frac 12 {\mathsf E}|X-X'|,
\end{equation*}
where \ $X$ \ and \ $X'$ \ are independent random variables with distribution \ $F$ \ and finite first moment. Both LogS and CRPS are negatively oriented, that is smaller score values indicate better predictive performance.

For a given lead time the goodness of fit of competing TCC forecasts in terms of probability distributions are compared with the help of the mean CRPS and mean LogS values \ $\overline{\crps}$ \ and  \ $\overline{\logs}$, \ respectively, over all forecast cases in the verification data. Further, the improvement in CRPS and LogS with respect to a reference model can be quantified using the continuous ranked probability skill score (CRPSS) and logarithmic skill score (LogSS), respectively, defined as
\begin{equation*}
 \crpss := 1- \frac{\overline{\crps}}{\overline{\crps}_{ref}} \qquad \text{and} \qquad \logss := 1- \frac{\overline{\logs}}{\overline{\logs}_{ref}},
\end{equation*}
where \ $\overline{\crps}_{ref}$ \ and  \ $\overline{\logs}_{ref}$ \ denote the mean CRPS and  LogS of the reference approach  \citep[see e.g.][]{murphy73,gr07}. Note that both CRPSS and LogSS are positively oriented, that is larger skill scores mean better predictive performance. 

Further, following the suggestions of \citet{gr11}, statistical significance of the differences between the verification scores is examined using the Diebold-Mariano \citep[DM;][]{dm95} test, which allows accounting for the temporal dependencies in the forecast errors. In simultaneous testing for the different stations we also address spatial dependencies by applying a \citet{bh95} algorithm to control the false discovery rate at a $5\,\%$ level of significance \citep[see e.g.][]{wilks16}. We further provide confidence intervals for mean score values and skill scores, which are obtained with the help of $2\,000$ block bootstrap samples using the stationary bootstrap scheme with mean block length determined according to \citet{pr94}.

Finally, a simple tool of visual perception of calibration is the probability integral transform (PIT) histogram, where the PIT is defined as the value of the predictive cumulative distribution (CDF) at the validating observation, with a possible randomization at points of discontinuity \citep{gr13}. In the case of proper calibration PIT should follow a uniform distribution on the \ $[0, 1]$ \ interval, moreover, if uniformity fails to be achieved, the shape of the PIT histogram provides information about the possible reason of the problem.

\section{Results}
  \label{sec4}
  
All calibration approaches presented in Section \ref{sec3} require training data which should be large enough to provide numerical stability and reasonable predictive performance. Following \citet{hspbh14}, we here focus on local calibration, i.e., post-processing of forecasts for a given station is performed using only training data of that particular station. Therefore, relatively long training periods are required to achieve a suitably large training set. 
In order to ensure comparability with the reference approaches we consider 5-year training periods and both non-seasonal and seasonal training schemes as in \citet{hhp16}. In the non-seasonal training, forecasts and observations of 5 calendar years (e.g., 1 January 2003 -- 31 December 2007) are used to train the model for calibration of TCC ensemble forecasts for the whole next calendar year (1 January -- 31 December 2008), then the training period is rolled ahead by one year (1 January 2004 -- 31 December 2008). In the seasonal approach, two different seasons are considered covering April -- September and October -- March, and TCC ensemble forecast for a given day is calibrated using training data from the same season only. The use of 5-year training periods means that predictive PMFs are available for the time interval between 1 January 2007 and 20 March 2014 (2636 calendar days), where one can test the forecast skill of the post-processing methods presented in Section \ref{sec3}.

Further, as suggested by \citet{hhp16}, numerical problems with LogS calculation are avoided by replacing  unrealistically low values \ $p_F(y_j)$ \ of the predictive PMF corresponding to okta \ $y_j$ \ with a probability \ $p_{\min}$ \ ensuring that with a \ $1\%$ \ chance one observes okta \ $y_j$ \ at least once during the training period. Translated to formulae, this means that instead of \ $p_F(y_j)$ \ we consider \ $\max\big\{p_{\min},p_F(y_j)\big\}$, \ where \ $p_{\min}$ \ solves \ $0.01=1-\big(1-p_{\min} \big)^{T}$ \ with \ $T$ \ being the length of the training period in days, and adjust the probabilities to get a PMF again \citep[for more details see][]{hhp16}. Note that this is only a minor technical adjustment and compared with the original predictive PMFs it results in negligible differences in CRPS or PIT values.

\subsection{Implementation details}
\label{subs4.1-impl}
Here, we discuss implementation details for the different statistical and machine learning methods for TCC post-processing.

\subsubsection*{Multiclass and proportional odds logistic regression}

Both models have several implementations. Here, coefficients of the various MLR and POLR models are estimated with the help of {\tt R} packages {\tt nnet} and {\tt MASS} \citep{vr02}, respectively. Note that the implementation based on the \texttt{nnet} package utilizes neural networks for estimating the parametric MLR model \eqref{eq:mlr} which is a fundamentally different use of neural networks compared to our MLP models introduced in Section \ref{subs3.1}.

\subsubsection*{Multilayer perceptron neural networks}

In our computations we apply the {\tt patternnet} function of \texttt{Matlab} with two hidden layers, consisting of 10 and 15 neurons. Both hidden layers use the hyperbolic tangent as activation function. We consider the  LogS as loss function (sometimes termed cross-entropy in the machine learning literature) with a $0.1$ regularization parameter and scaled conjugate gradient as minimization algorithm. In each 5-year training period (both for the seasonal and non-seasonal approaches) the corresponding data set is split into a training and validation set, the latter is a randomly selected subset consisting of $15\,\%$ of the data. As an alternative to the 5-year rolling training period, training with a growing data set using all available forecast cases from the previous years and simultaneously increasing the weight of the regularization term was also tested. However, this approach did not result in an improved forecast skill.

\subsubsection*{Random forests}

Our implementation of RF models is based on the \texttt{R} package \texttt{XGBoost} \citep{xgboostpckg}. The tuning parameters (depth of trees, number of predictors sub-sampled at each splitting node) for a specific observation station and forecast horizon are determined as follows. The first of the rolling 5-year training periods consisting of the years 2002--2006 is split into an initial training set (years 2002--2005) and a validation set (year 2006). For all combinations of tree depths between 2 and 4, and numbers of predictors between 1 and 3, RF models consisting of 300 trees are estimated based on the initial training set, and evaluated on the validation set using the LogS. The combination of tuning parameters resulting in the lowest LogS on the validation set is then used to fit a RF model consisting of 1000 trees for the full training set (years 2002--2005), and to produce forecasts for the first out-of-sample test set (year 2007). To limit computational costs, this optimal combination of tuning parameters is also used for all subsequent 5-year training periods for that specific station and lead time.  

For rolling 5-year training periods, tree depths of 2, 3 and 4 are selected in around 43\,\%, 36\,\% and 21\,\% of the cases, respectively. The chosen number of predictors for subsampling is slightly more evenly distributed, and the most frequently selected tuning parameter combination consists of trees of depth 3 with 3 predictors sub-sampled at each split (around 17\,\% of all cases). Note that since initial tests did not indicate improvements in predictive performance and RF models often tend to be relatively robust to the choice of tuning parameters, we did not consider a more extensive set of possible parameter values in order to limit computational costs.

\subsubsection*{Gradient boosting machines}

We implement GBM models based on the \texttt{R} package \texttt{XGBoost} \citep{xgboostpckg}. Throughout, we use shrinkage with a learning rate of \ $\lambda = 0.1$ \ which reduces the influence of each individual tree \ $h_m^c$ \ by adding a scaled version of that tree only. To further prevent overfitting, we determine the number of boosting iterations \ $M$ \ for a fixed tree depth by using an early stopping criterion. To that end, each 5-year training set is split into an initial training set (first 4 years), and a validation set (last year). GBM models of the form \eqref{eq:gbm} are then estimated iteratively for \ $m = 1,2, \ldots $ \ based on the initial training set until the LogS on the validation set has not improved during the last 25 iterations. This process is repeated for all tree depth values between 1 and 4, and the combination of tree depth and corresponding optimal number of boosting iterations that results in the best LogS on the training set is selected as set of tuning parameters. The final out-of-sample forecasts for the test set are produced based on a GBM model fitted on the full training set using these tuning parameters. A separate set of tuning parameters is determined according to the procedure described above for any combination of station and lead time, and any of the rolling 5-year training periods. 

For models with a rolling 5-year training period an optimal tree depth of 1 is selected for around 86.5\,\% of all GBM models, a depth of 2 in around 11.5\,\% of the cases and a depth of 3 or 4 in less than 2\,\%. The average number of boosting iterations is 78.3, but generally depends on the corresponding tree depth.

 The procedures to determine optimal tuning parameters of RF and GBM models described above are applied separately to the two seasons when fitting seasonal RF and GBM models. Therefore, the sets of optimal tuning parameters differ not only by station, lead time and year (only for GBM), but also by season for those variants.

\subsection{Post-processing of TCC ensemble forecasts}
  \label{subs4.1}

\begin{figure}[t]
\begin{center}
\epsfig{file=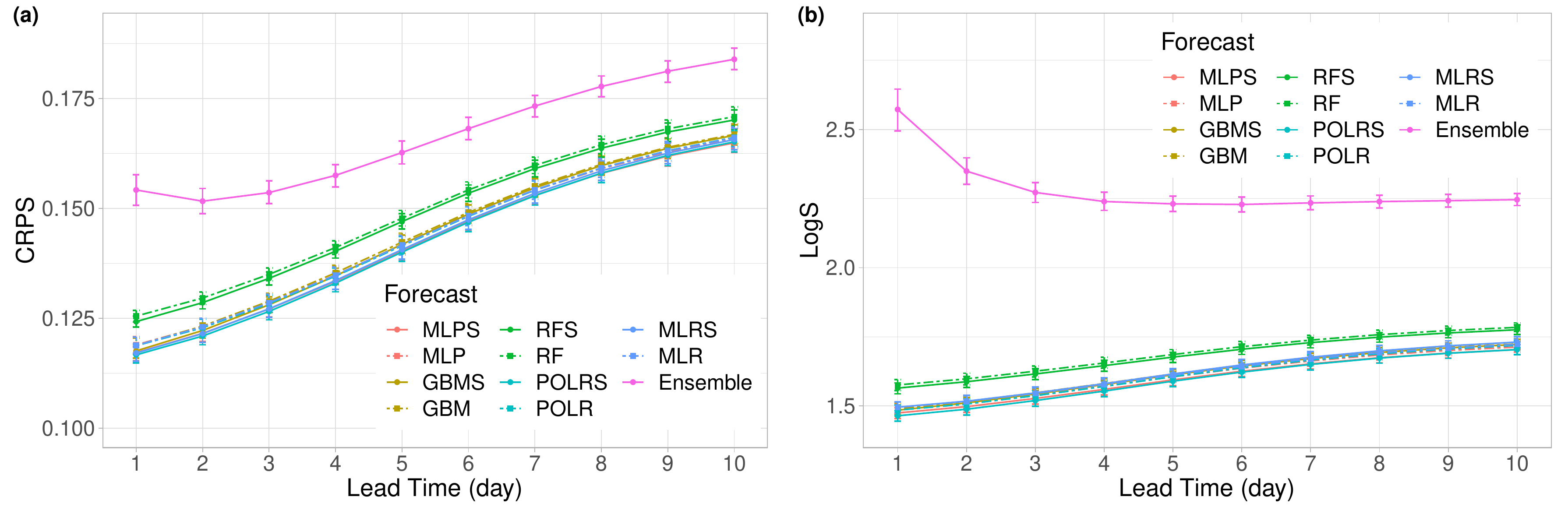, width=\textwidth}
\end{center}
\caption{Mean CRPS (a) and LogS (b) of the raw ensemble and post-processed forecasts together with $95\,\%$ confidence intervals.}
\label{fig:crps_logs}
\end{figure}

As a first step we investigate the post-processing of TCC ensemble forecasts using the MLP, RF and GBM approaches. As references we consider the raw TCC ensemble forecast and the MLR and POLR models. All calibrated forecasts are based on the 7-dimensional feature vector \ $\big(\overline f_{\text{ENS}}, f_{\text{CTRL}}, f_{\text{HRES}}, s^2, p_0, p_1, I\big)^{\top}$ \ except the MLR, where following \citet{hhp16} the number of parameters is reduced by omitting the interaction term \ $I$. \ Note that the MLP model was also tested with the 52-member TCC forecast ensemble as feature vector, however, this approach did not result in an improved predictive performance.  Further, following again the suggestions of \citet{hhp16}, in the POLR model the coefficients of \ $\overline f_{\text{ENS}}, \ f_{\text{CTRL}}$ \ and \ $f_{\text{HRES}}$ \ are forced to be non-negative by iterative exclusion of covariates with negative weights. Finally, for all five calibration methods we test both non-seasonal and seasonal training, forecasts obtained using the latter are referred as MLPS, RFS, GBMS, MLRS and POLRS, respectively.

\begin{figure}[t]
\begin{center}
\epsfig{file=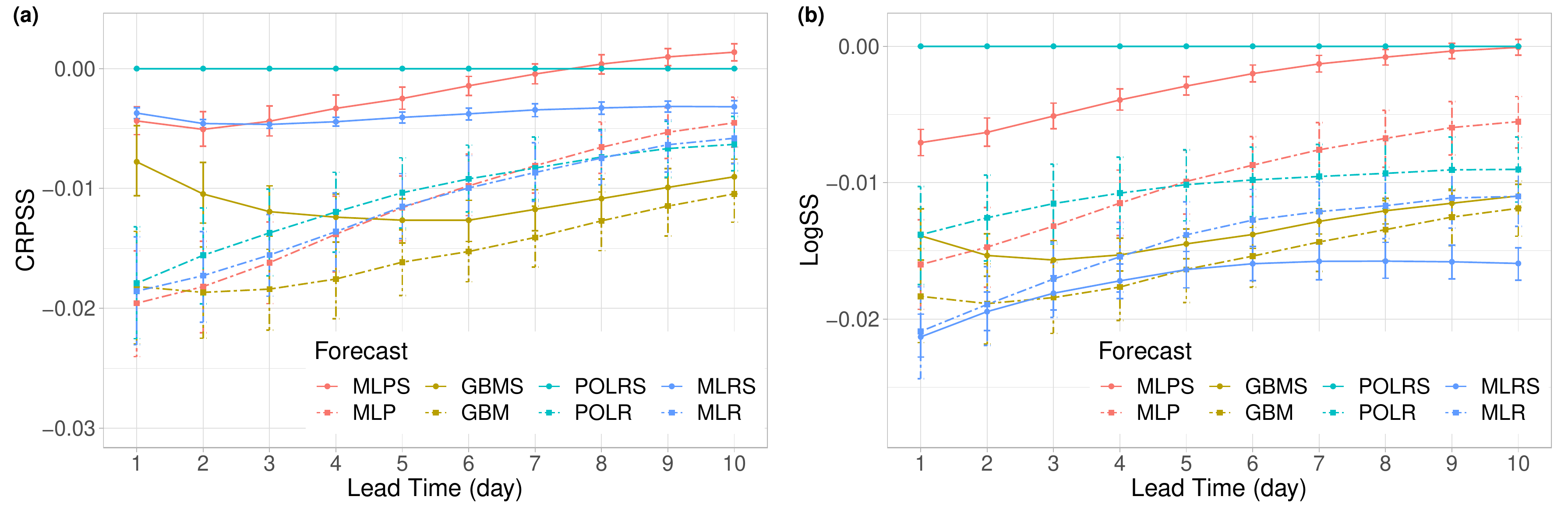, width=\textwidth}
\end{center}
\caption{CRPSS (a) and LogSS (b) with respect to the POLRS model of MLPS, MLP, GBMS, GBM, MLRS, MLR and POLR forecasts together with $95\,\%$ confidence intervals.}
\label{fig:crpss_logss}
\end{figure}

Figure \ref{fig:crps_logs} shows the mean CRPS and LogS of the raw ensemble and post-processed forecasts together with $95\,\%$ confidence intervals as functions of the lead time. All calibrated TCC forecasts outperform the raw ensemble by a wide margin and one can observe a clear grouping of the various approaches. The first group, resulting in the lowest mean CRPS and LogS values, consists of the MLP, GBM, POLR and MLR methods and their seasonally estimated versions showing very small differences in forecast skill. The second group contains the non-seasonally and seasonally estimated RF forecasts, where the latter results in slightly lower score values than the former.

\begin{figure}[t]
\begin{center}
\epsfig{file=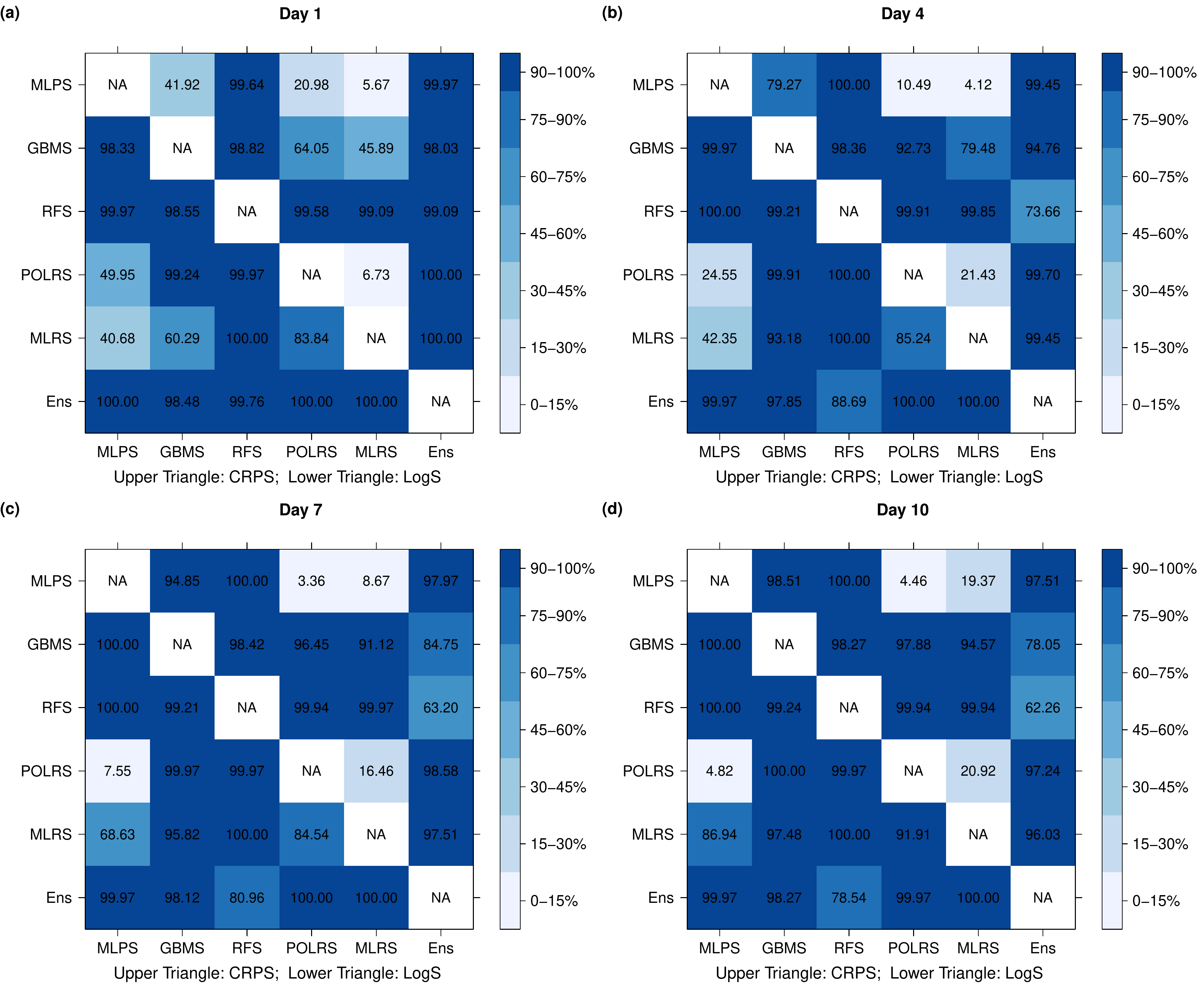, width=\textwidth}
\end{center}
\caption{Proportion of stations with significantly different mean CRPS (upper triangle) and LogS (lower triangle) at a $5\,\%$ level of significance for lead times 1 (a), 4 (b), 7 (c) and 10 (d) days.}
\label{fig:dmtest}
\end{figure}

One can compare the performance of the forecasts in the first group more easily by examining Figure \ref{fig:crpss_logss}, where the CRPSS and LogSS  with respect to the POLRS forecasts are plotted, which showed the best forecast skill among the methods studied in \citet{hhp16}. According to Figure \ref{fig:crpss_logss}(a), in terms of the mean CRPS, POLRS outperforms its competitors up to day 7, whereas for longer lead times MLPS has the best predictive performance. In general, forecasts based on seasonal training result in lower mean CRPS than their non-seasonal counterparts, however, the differences decrease with the increase of the lead time. Results in terms of the LogS shown in Figure \ref{fig:crpss_logss}(b) indicate a different behavior and ranking of the models in that the mean LogS of the MLPS approach reaches that of the POLRS model only at day 10 and MLRS underperforms all other methods for all lead times.

\begin{figure}[h!]
\begin{center}
\epsfig{file=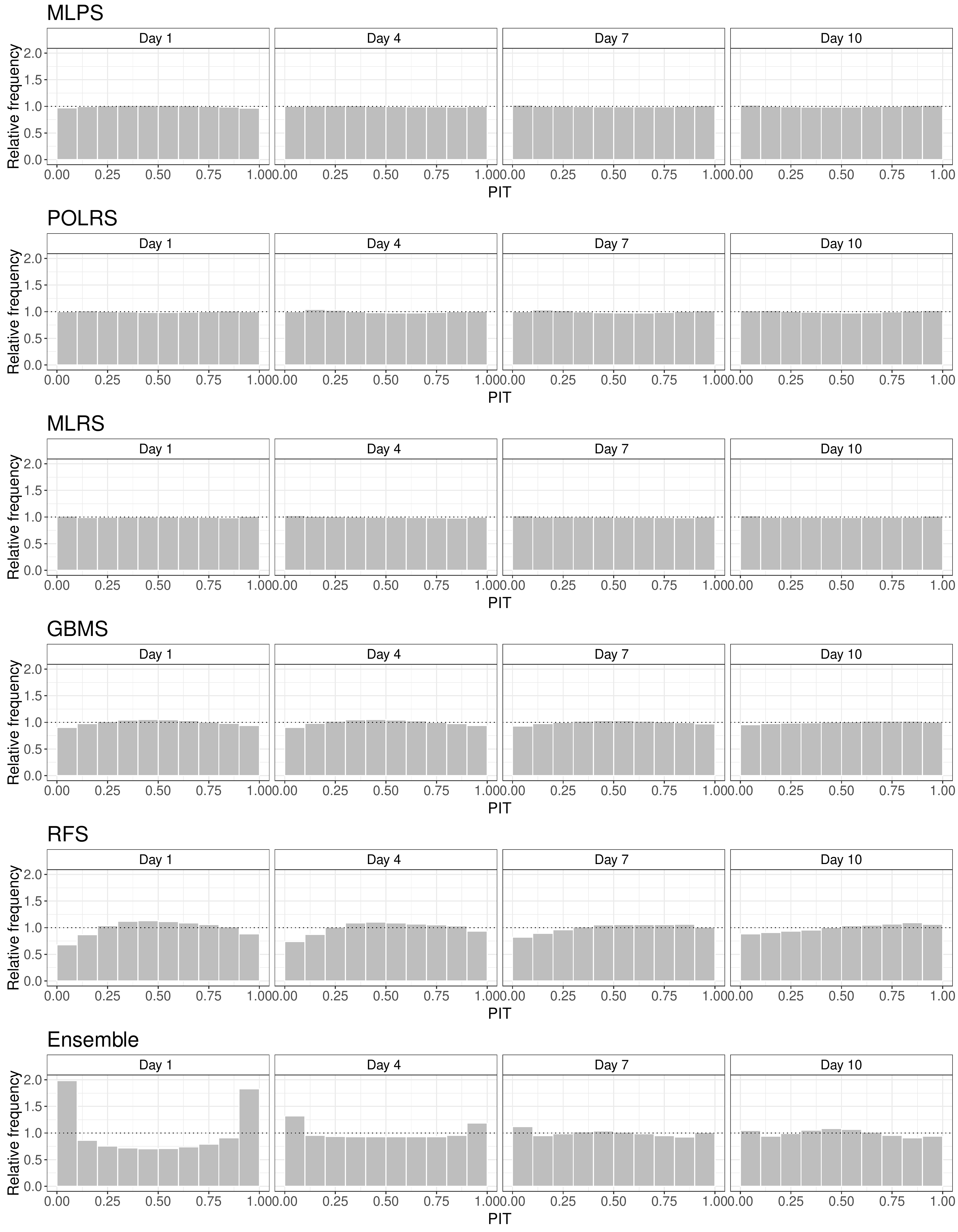, width=.95\textwidth}
\end{center}
\caption{PIT histograms over all stations and dates (3300 stations, 2636 days) of the seasonally trained calibration approaches and the raw ensemble at days 1, 4, 7 and 10.}
\label{fig:pit}
\end{figure}

These observations are further supported by Figure \ref{fig:dmtest} showing the proportion of stations where DM test indicates significant difference in mean CRPS and LogS for lead times 1, 4, 7 and 10 days. To simplify the presentation here we compare just the raw ensemble and the seasonally trained versions of the calibration approaches, as in general seasonal models outperform their non-seasonal counterparts. Raw ensemble and RFS forecasts are clearly separated from the other four approaches for all lead times, as most entries of the corresponding cells are close to $100\,\%$. For longer lead times GBMS also differs significantly from its competitors in almost all stations both in terms of CRPS and LogS. On contrary, the increase of the lead time reduces the proportion of stations where the mean LogS of MLPS and POLRS forecast differ, whereas in terms of the mean CRPS after decrease one can observe a slight increase. This behaviour is in line with the MLPS skill scores of Figures \ref{fig:crpss_logss}(b) an \ref{fig:crpss_logss}(a), respectively. Overall, we note that even though the absolute differences in terms of CRPS and LogS between the different methods are relatively small, they thus are often statistically significant for a large proportion of the stations. 

The positive effect of post-processing can also be observed in the PIT histograms in Figure \ref{fig:pit}, where again, only the results for better performing seasonally trained models are shown. The U-shaped histograms of the raw ensemble at days 1 and 4 clearly indicate underdispersion, whereas at days 7 and 10 a small hump starts to appear. RFS forecasts are overdispersive for short lead times, and develop some bias as the forecast horizon increases. GBMS forecasts exhibit the same behaviour, however, to a much smaller extent. The PIT histograms of POLRS and MLPS are almost perfectly flat, indicating a better calibration compared to the other methods.

\subsection{Post-processing using an extended feature set}
\label{subs4.2}

\begin{figure}[t]
\begin{center}
\epsfig{file=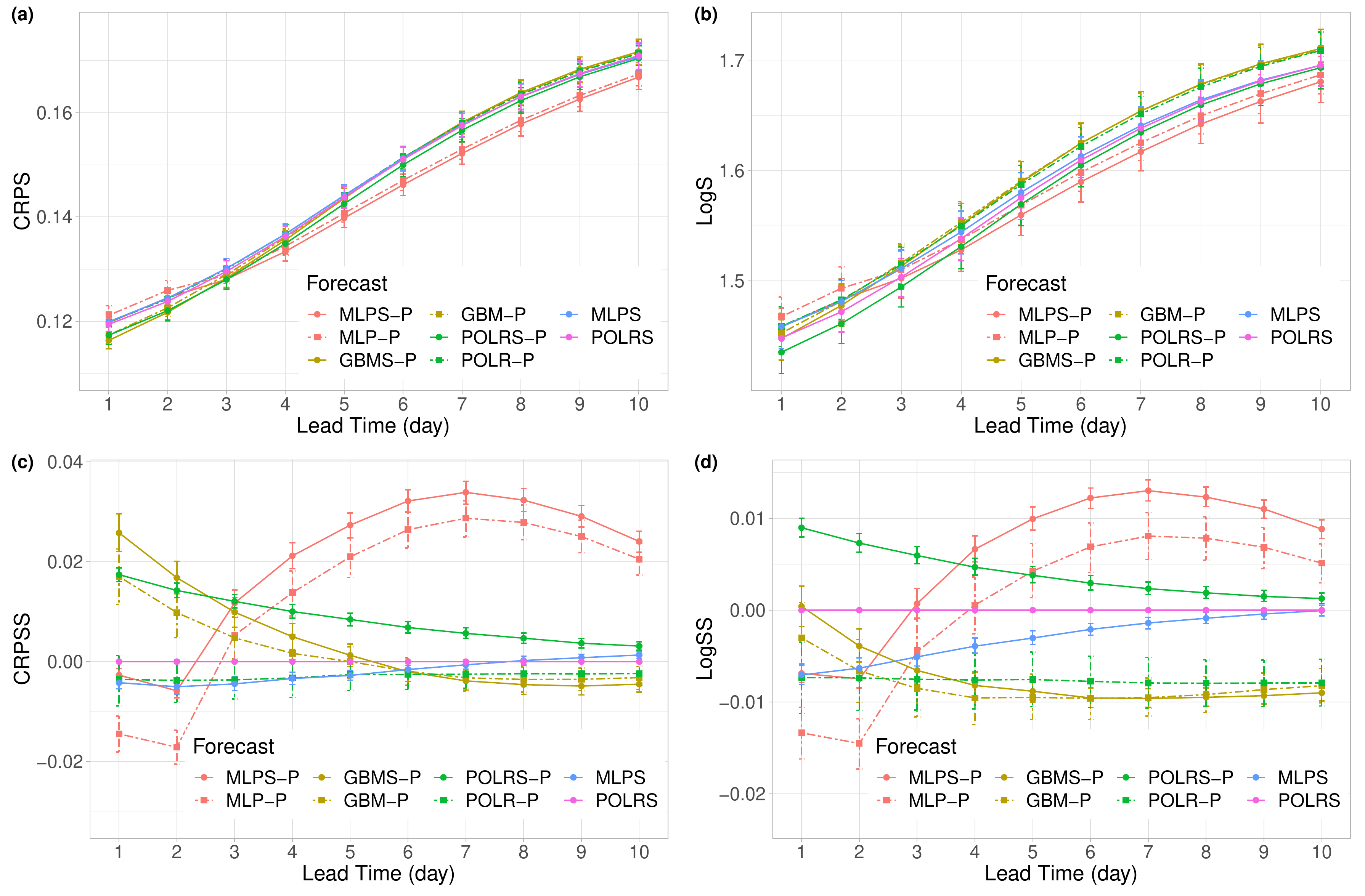, width=\textwidth}
\end{center}
\caption{CRPS (a) and LogS (b) of different MLP, GBM and POLR forecasts and the corresponding skill scores with respect to the POLRS model (c,d) together with $95\,\%$ confidence intervals.}
\label{fig:crps_logs_prec}
\end{figure}

The added value of incorporating additional features based on geographical data of SYNOP stations and/or forecasts of other weather variables has been demonstrated in various recent articles on post-processing \citep[e.g.,][]{tmzn16,rl18,bwks19}. Due to the direct connection to clouds \citep{mishra19}, functionals of precipitation ensemble forecasts  represent a natural choice for additional predictors. We here use the mean \ $\overline f_{\text{PREC}}$ \ of the ECMWF 51-member precipitation forecast as additional covariate and investigate the performance of MLP, GBM and POLR approaches, showing the best forecast skill in Section \ref{subs4.1}, with extended feature vector \ $\big(\overline f_{\text{ENS}}, f_{\text{CTRL}}, f_{\text{HRES}}, s^2, p_0, p_1, I, \overline f_{\text{PREC}}\big)^{\top}$. \ Again, we consider both non-seasonal and seasonal training, the corresponding models are referred as MLP-P, GBM-P, POLR-P and MLPS-P, GBMS-P, POLRS-P, respectively. 

\begin{figure}[t]
\begin{center}
\epsfig{file=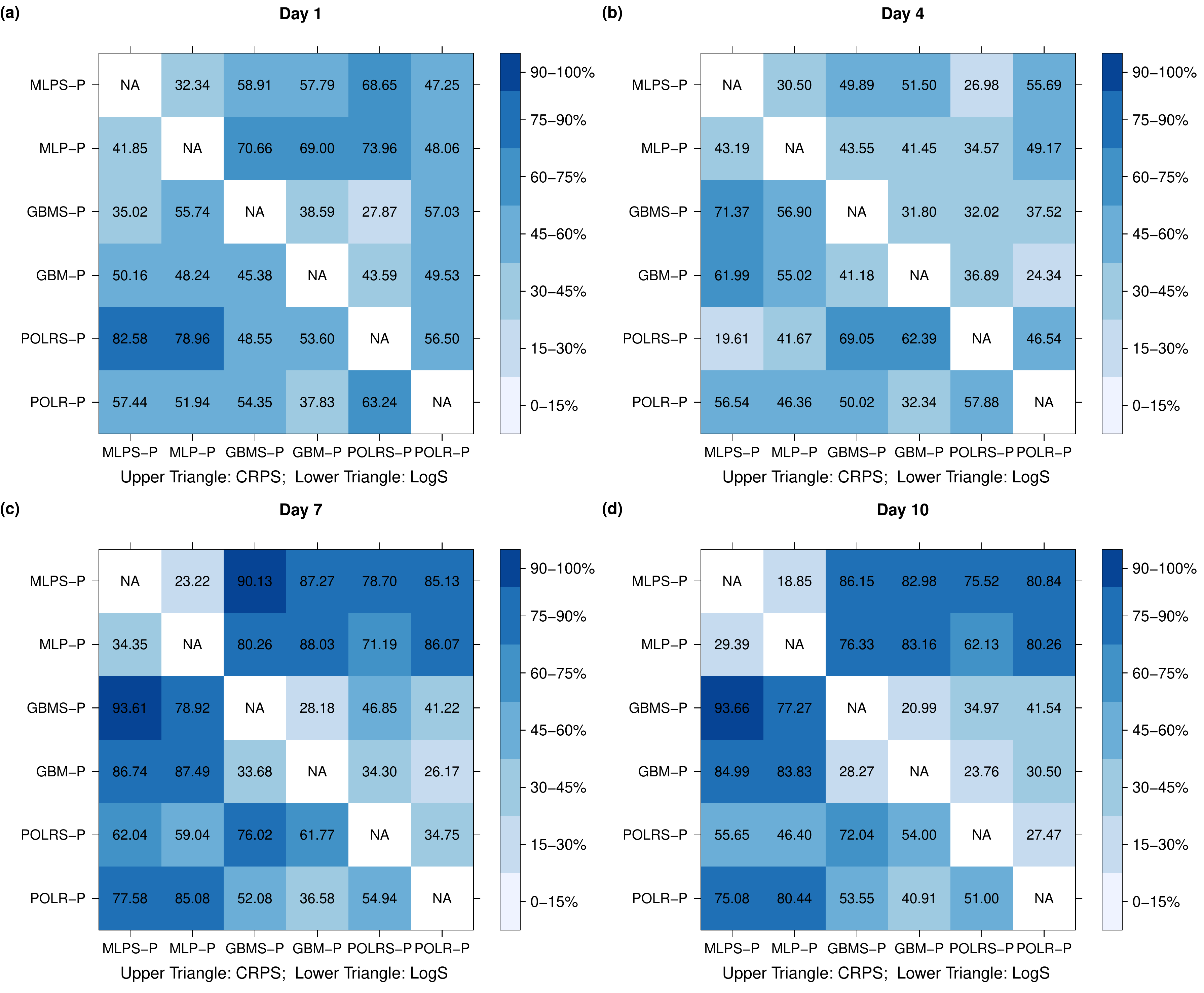, width=\textwidth}
\end{center}
\caption{Proportion of stations with significantly different mean CRPS (upper triangle) and LogS (lower triangle)  at a $5\,\%$ level of significance for lead times 1 (a), 4 (b), 7 (c) and 10 (d) days.}
\label{fig:dmtest_prec}
\end{figure}

According to  Figures \ref{fig:crps_logs_prec}(a) and \ref{fig:crps_logs_prec}(b), where the mean CRPS and LogS values of different MLP, GBM and POLR forecasts are plotted as functions of the lead time, and Figures \ref{fig:crps_logs_prec}(c) and \ref{fig:crps_logs_prec}(d) showing the corresponding skill scores with respect to the POLRS model, the additional covariate results in different effects for the MLP, and the GBM and POLR models.  After day 2 MLP models using also precipitation forecasts significantly outperform MLP models based on TCC forecasts only in terms of both CRPS and LogS regardless of the training scheme (MLP is not shown), moreover, for longer lead times MLPS-P and MLP-P show the best predictive performance. In contrast, the use of precipitation has the highest effect on POLR models at day 1 and the differences between POLRS-P and POLRS and POLR-P and POLR models (POLR is not shown) are decreasing with the increase of the lead time. The same phenomenon can be observed with GBMS and GBM models (not shown). The use of precipitation forecast substantially improves the predictive performance, however, the difference decreases with the increase of the lead time. Up to day 5 GBMS-P and GBM-P approaches result in lower mean CRPS than the POLRS model, whereas for days 1 and 2 GBMS-P outperforms POLRS-P and MLPS-P.

\begin{figure}[t!]
\begin{center}
\epsfig{file=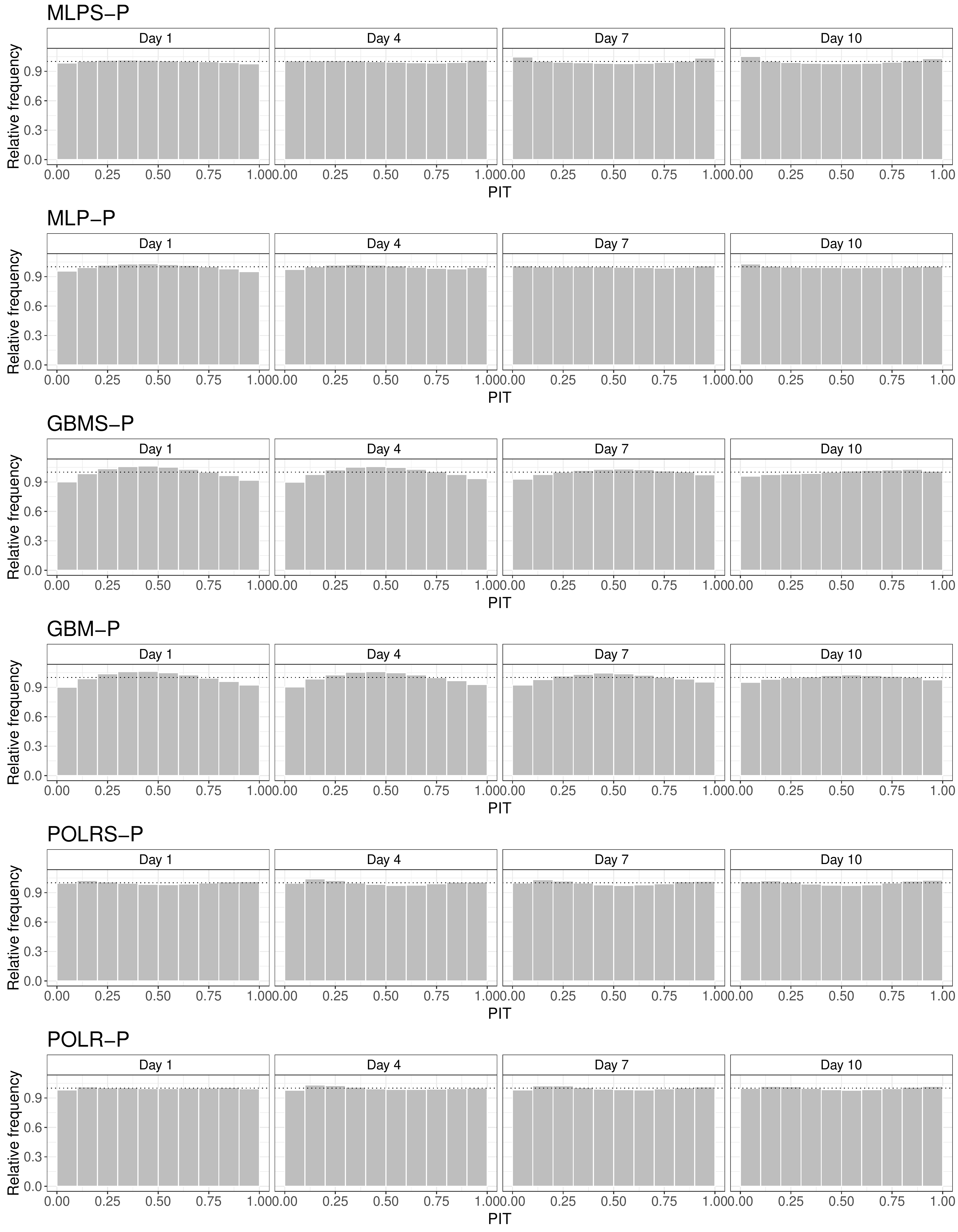, width=.95\textwidth}
\end{center}
\caption{PIT histograms over all stations and dates (2239 stations, 2636 days) of the calibration approaches using precipitation forecasts at days 1, 4, 7 and 10.}
\label{fig:pit_prec}
\end{figure}

These results are in line with proportions of stations with significantly different mean CRPS and LogS values provided in Figure \ref{fig:dmtest_prec}, where we consider only the models with the extended feature set in the interest of visual clarity. For instance, the proportion of stations where the mean CRPS of GBMS-P and GBM-P models differ shows a monotone decreasing sequence of $38.59\,\%, \ 31.80\,\%, \ 28.18\,\% \ 20.99\,\%$, mimicking the decreasing distance of the corresponding curves in Figure \ref{fig:crps_logs_prec}(c), while the bow of the CRPSS of MLPS-P with respect to POLRS and the decrease of the CRPSS of POLRS-P matches the change of the corresponding entries ($68.65\,\%, \ 26.98\,\%, \ 78.70\,\%, \ 75.72\,\%$) in Figure \ref{fig:dmtest_prec}.

Addressing calibration, Figure \ref{fig:pit_prec} shows the PIT histograms of the calibration approaches using precipitation forecasts at days 1, 4, 7 and 10. In general, all six methods result in rather well calibrated predictive PMFs for all lead times. The histograms of GBMS-P and GBM-P approaches are overdispersive for all lead times, whereas MLPS-P, MLP-P and POLR-P are slightly overconfident only at day 1, which transforms to a small underdispersion at longer lead times. Note that in contrast to Figure \ref{fig:pit}, which is based on verification data of 3330 locations, here we consider PIT values for just 2239 SYNOP stations where precipitation ensemble forecasts are also available. However, this reduction does not change the general shape of the PIT histograms of the raw ensemble and the MLPS, GBMS and POLRS forecasts, so they are not shown in this case. Finally, the general behaviour of the MLPS, MLP, GBMS, GBM, POLRS and POLR forecasts in terms of PIT values is almost completely inherited to the corresponding MLPS-P, MLP-P, GBMS-P, GBM-P, POLRS-P and POLR-P approaches.

\section{Discussion}
\label{sec5}

We investigate various machine learning classifiers for statistical post-processing of total cloud cover ensemble forecasts. In particular, we consider multilayer perceptron neural networks, random forest methods and gradient boosting machines, which are tested on ECMWF global TCC ensemble forecasts with lead times of $1,2, \ldots ,10$ days and the corresponding discrete SYNOP observations. Raw TCC ensemble forecasts, multiclass and proportional odds logistic regression are used as reference models, and we consider both seasonal and non-seasonal training \citep[following][]{hhp16}.

First we investigate the settings of \citet{hhp16}, where the classification is based on predictors calculated from the TCC ensemble forecasts only. In general, all post-processing methods significantly outperform the raw ensemble for all lead times both in term of the mean CRPS and the mean LogS over the verification data, and the corresponding PIT histograms are closer to the uniform distribution than those of the raw forecasts. Seasonally trained models further result in slightly better predictive performance than their non-seasonal counterparts. RF models underperform their competitors, whereas the difference between MLP, GBM, POLR and MLR approaches are generally small. For short and medium forecast horizons the POLR model with seasonal training occurs to be the most skillful, closely followed by the seasonally trained MLP model which performs best for long lead times. Several of the probabilistic classification methods exhibit complementary systematic errors in calibration. Therefore, forecast combination techniques along the lines of \citet{bl18} could potentially improve predictive performance. The related topic of calibrating and combining probabilistic classifiers has recently received some interest in the machine learning literature, see e.g.\ \citet{kull19}.

Due to the flexibility of neural network model architectures, particularly the MLP model provides several promising starting points for future extensions. For example,  long short-term memory neural networks \citep{hr97} are widely used for time series modeling and may allow to incorporate temporal dependencies of forecast errors of the raw ensemble predictions. Further, techniques along the lines of station embeddings proposed in \citet{rl18} could potentially help construct a single MLP model jointly for all stations which still is locally adaptive. 

The use of the mean precipitation accumulation as additional covariate further improves the predictive performance and changes the ranking of the different methods. With this extended feature set the seasonal POLR model exhibits the best overall performance only for short lead times, after days 3 -- 4 it is significantly outperformed both by the seasonally and non-seasonally trained MLP. However, in general, the advantage of the extended set of covariates fades with the increase of the lead time.

The improved performance when information on precipitation is added further indicates advantages of modern machine learning methods such as GBM and MLP for total cloud cover prediction. By contrast to the classical MLR and POLR approaches, these methods allow to add additional predictors in a straightforward manner and provide tools for avoiding overfitting. The inclusion of further predictor variables such as, for example, indices of atmospheric stability, pressure, humidity and temperature information at upper levels of the atmosphere, or seasonal information may further improve predictive performance. Further, more complex machine learning models incorporating many predictors may not only improve TCC predictions, but may also allow to better understand the shortcomings of the raw ensemble predictions utilizing techniques such as measures of feature importance \citep{breiman01,rl18}.

\bigskip
\noindent
{\bf Acknowledgments.} \  Essential part of this work was made during the visit
of S\'andor Baran at the Heidelberg Institute of Theoretical Studies.
S\'andor Baran further received support from the National Research, Development and Innovation Office under Grant No. NN125679. \'Agnes Baran and S\'andor Baran were supported by the EFOP-3.6.2-16-2017-00015 project. The project was co-financed by the Hungarian Government and the European Social Fund. Sebastian Lerch was further supported by the Deutsche Forschungsgemeinschaft through SFB/TRR 165 ``Waves to Weather''. The authors are grateful to Stephan Hemri for providing the data and for useful suggestions and comments.

\end{document}